\documentclass[conference]{IEEEtran}
\usepackage{times}

\usepackage[numbers]{natbib}
\usepackage{multicol}
\usepackage[bookmarks=true]{hyperref}
\usepackage{amsmath}
\usepackage{amsfonts}
\usepackage{amssymb}
\usepackage{booktabs}
\usepackage{subcaption}
\usepackage{graphicx}
\usepackage{pifont}
\graphicspath{{./figs/}}

\begin{document}

\title{Learning Probabilistic Intersection Traffic Models for Trajectory Prediction }


\author{\authorblockN{Andrew Patterson, Aditya Gahlawat and Naira Hovakimyan}
\authorblockA{ Department of Mechanical Science and Engineering \\
University of Illinois Champaign-Urbana \\
mail: appatte2@illinois.edu }}

\maketitle

\begin{abstract}
Autonomous agents must be able to safely interact with other vehicles to integrate into urban environments. The safety of these agents is dependent on their ability to predict collisions with other vehicles' future trajectories for replanning and collision avoidance. The information needed to predict collisions can be learned from previously observed vehicle trajectories in a specific environment, generating a traffic model. The learned traffic model can then be incorporated as prior knowledge into any trajectory estimation method being used in this environment.
This work presents a Gaussian process based probabilistic traffic model that is used to quantify vehicle behaviors in an intersection. The Gaussian process model provides estimates for  the average vehicle trajectory, while also capturing the variance between the different paths a vehicle may take in the intersection.
The method is demonstrated on a set of time-series position trajectories. 
These trajectories are reconstructed by removing object recognition errors and missed frames that may occur due to data source processing. 
To create the intersection traffic model, the reconstructed trajectories are clustered based on their source and destination lanes. 
For each cluster, a Gaussian process model is created to capture the average behavior and the variance of the cluster. 
To show the applicability of the Gaussian model, the test trajectories are classified with only partial observations. 
Performance is quantified by the number of observations required to correctly classify the vehicle trajectory. 
Both the intersection traffic modeling computations and the classification procedure are timed. 
These times are presented as results and demonstrate that the model can be constructed in a reasonable amount of time and the classification procedure can be used for online applications.
\end{abstract}

\IEEEpeerreviewmaketitle

\section{Introduction}
\label{sec:introduction}
Technology projections anticipate an unprecedented level of human-robot interaction in the coming decades, especially with autonomous vehicles interacting with pedestrians and human-operated vehicles. 
Interaction between humans and robots present unique challenges: heterogeneous communication methods, asymmetric capabilities, and non-cooperative behaviors. 
These challenges limit an autonomous system's ability to interact with human agents since the future behavior of these agents is unknown and possibly time-varying~\cite{yoon_socially_2019}. 
If safe interaction cannot be assured, even for non-communicative agents, their applications will be limited. To assure safety, methods of estimating the future behavior of a vehicle must, therefore, be fast and provide an accurate representation of the variability of human-operated vehicles.

Consider an autonomous vehicle in an intersection. When vehicles are cooperative, they can directly employ path replanning methods to negotiate a collision-free trajectory, as described in~\cite{kaminer17} and~\cite{bilal19}.  Without this communication, feedback methods can be used that guarantee collision avoidance,~\cite{marinho18} and~\cite{cichella17}. However, the assumptions used in these methods are conservative and often require sacrificing performance objectives. Striking a balance between these methods, without an exact model for other vehicles' future behavior, can be achieved if the uncertain future trajectories can be estimated. Given these trajectory estimates, collisions can be predicted by finding intersections between a desired trajectory and another vehicle's estimated trajectory. In situations with limited communication, collision prediction is necessary for safe interaction and cooperation. Combining replanning methods, such as those found in~\cite{wurts18} and~\cite{paden16}, with collision prediction methods can allow autonomous vehicles to safely complete their objectives, even without the knowledge of other vehicles' exact trajectories.

Trajectory estimation methods are central to safe interaction as they allow safe replanning. Though these methods predict the future trajectory of a vehicle based on recent observations of that vehicle's movement, incorporating additional information can help increase the accuracy of these methods. 
Using only observations of a vehicle's past motion to construct a time-series forecast may lead to an explosion in uncertainty that overestimates the possibilities in the future. 
To solve this problem, additional information in the form of an intention can be included. 
The intention provides additional information that can reduce uncertainty estimates and indicate how the vehicle's behavior will change. Intention based methods can be found in~\cite{patterson2019},~\cite{xin_intention-aware_2018} and~\cite{yokoyama18}. However, to use an intention it must be first estimated.
One additional source of information is the behavior of similar vehicles in the same location. This information enables the construction of informative prior assumptions for specific scenarios. For example, if every car in the past has turned left out of a lane, one could reasonably assume that the car we see in that lane now will also turn left. This information can be readily used to calculate intention.

Some methods attempt to learn the future trajectories through deep learning architectures, without explicitly modeling intentions. These methods train on a large quantity of data and as a result require significant computing time for training. The authors of~\cite{morton_analysis_2017} use a recurrent neural network for trajectory estimation. The authors of~\cite{pan_lane_2019} include the vehicle's intention implicitly as part of an aggregated hidden state. The authors of~\cite{xin_intention-aware_2018} use hierarchical long short term memory networks. The first network identifies the vehicle's intention, and the second one predicts the future trajectory.  These methods have prediction horizons less than 5~seconds and produce only the expected trajectory.

Rather than just estimating the expected future trajectory, some methods create a probabilistic model for the locations where a vehicle may be in the future.  Probabilistic models attempt to quantify modeling errors so that their effects can be incorporated into the collision avoidance and replanning decisions. A probabilistic traffic model allows the behavior of vehicles in a specific environment to be learned based on past observations. In these methods, incorporating additional information, such as the vehicle's intention, can help to increase the accuracy of these probabilistic estimates. The authors of~\cite{zhu_probabilistic_2019} compare five different probabilistic trajectory prediction methods and compare not only the average performance but also the variance of the performance. In these tests, they show that nonparametric regression methods are competitive with neural network-based methods in error prediction. The authors claim that this regression method has  a prediction horizon limited to less than 2~seconds due to scaling issues. However, the authors of~\cite{patterson2019} demonstrate the use of nonparametric regression for trajectory estimation with prediction horizons of 3~seconds and propose its use for fast online collision prediction. All these trajectory estimation methods  rely on a traffic model to predict trajectories. 

Methods of traffic modeling can be found in~\cite{ren2016}, which uses cubic spline basis functions for individual trajectories and estimates the probability of high-level actions by counting the number of cars that take a specific action. The authors of~\cite{barratt2019} focus on aerospace models using a Gaussian mixture model to represent distributions of take-off and landing trajectories, to create accurate airspace simulations.

\begin{figure}[t]
  \centering
  \includegraphics[width=0.85\columnwidth]{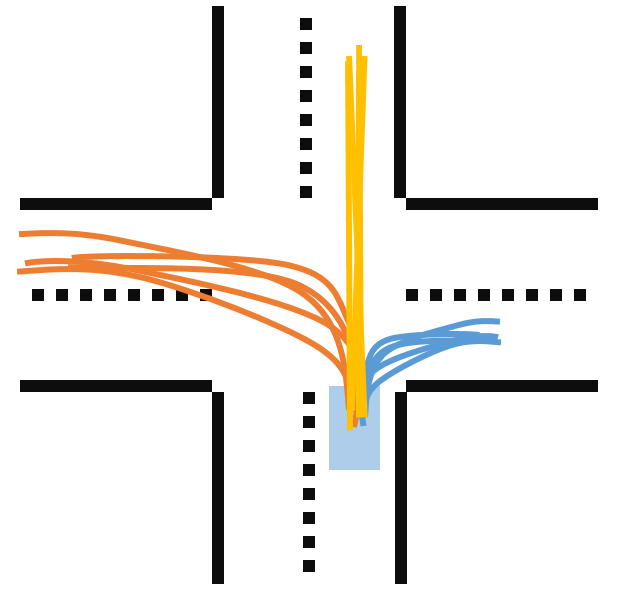}
  \caption{A probabilistic traffic model of the intersection shown above should capture both the average behavior of vehicles, as well as the observed deviation. In this figure, the car is shown as a blue rectangle and may turn right, left or continue straight. Possible trajectories are shown as colored lines, with variation within each set. }
  \label{fig:intro}
\end{figure}

The proposed method in this paper addresses the need for a vehicle behavior model that can be used directly in a trajectory estimation method or as an initial guess for training a more complex vehicle model.
We propose a vehicle behavior model, referred to as the traffic model. 
The model is probabilistic, capturing the variance in possible vehicle trajectories, which can be constructed quickly and is demonstrated in an online classification application.
The traffic model construction method is demonstrated in a four-way intersection, similar to the one shown in Figure~\ref{fig:intro}. An intersection is chosen for demonstration since it is a structured environment where vehicle trajectories are visually distinct. 
 This traffic model can be used for simulating different car trajectories or estimating the region of the intersection where cars are most likely to drive. 
 
The aforementioned  traffic model is generated offline from a set of observed trajectories. The trajectories are expected to be generated by a computer vision system observing online vehicle videos, such as those found in the companion dataset to~\cite{ren2016}. 
The  model is constructed using Gaussian process regression, a nonparametric regression method, explained in detail in~\cite{rasmussen06}. 
The intersection traffic is modeled such that each sample is a mean function of a vehicle trajectory. 
The variance of this Gaussian process is generated through an empirical variance on the means of the observed vehicle trajectories in the intersection. This intersection traffic model is designed for use in uncertainty estimation. 
Sampling the trajectory at a specific time will provide a Gaussian distribution in the x-y plane with the variance representing the variation in trajectories observed in this intersection. 
Once this traffic model is created, it can be used to classify an observed vehicle's behavior and forecast its trajectory based on the learned model, with the method presented in~\cite{patterson2019}. 
The proposed traffic model captures the uncertainty due to both measurement noise and different vehicle trajectories. We show that the proposed traffic model  can be generated quickly based on a set of 1000 simulated trajectories and that this model is suitable for online classification of an observed vehicle.   

\subsubsection{Contributions}
The paper presents (1) a probabilistic traffic modeling approach that can be used to characterize the mean and variance of vehicle trajectories, (2) a demonstration of the modeling algorithm in a four-way intersection, and (3) a demonstration of the model's use in the example application of online classification for partially observed vehicle trajectories. The probabilistic traffic model is designed to be included as prior knowledge in trajectory prediction applications for the safe operation of autonomous vehicles.  

In Section~\ref{sec:preliminaries}, an overview of mathematical concepts is presented, with topics including Gaussian process regression, the distance used for classification. In Section~\ref{sec:problem} we formalize the problem of learning the intersection traffic model. In Section~\ref{sec:method} we present our method, and in Section~\ref{sec:results} the results are presented. Finally, in Section~\ref{sec:conclusion} we conclude with an interpretation of the results and suggest future research directions.

\section{Preliminaries}
\label{sec:preliminaries}
In this section, we present the background used in both the traffic modeling and classification tasks. First, we present Gaussian processes and Gaussian process regression. This information is used in both the modeling and classification tasks. Secondly, we introduce two metrics used during the classification task. Finally, a brief description of the initial dataset is provided.
\subsection{Gaussian Processes}
Gaussian process regression is a data-based, nonparametric regression method. This work uses Gaussian process regression to create continuous time models from discretely observed trajectories. A general treatment of Gaussian process regression can be found in~\cite{rasmussen06}, and the specifics of time-series modeling can be found in~\cite{roberts12}.

A Gaussian process is a collection of random variables, where any finite numbers are jointly Gaussian. In this work, these processes are considered as a distribution over functions on $\mathbb{R}^d$ with
\begin{align*}
    F \sim \mathcal{GP}(M,K),
\end{align*}
where $M : \mathbb{R}^d \rightarrow \mathbb{R}$ is the mean function, and $K:\mathbb{R}^d \times \mathbb{R}^d \rightarrow \mathbb{R}$ is a symmetric positive definite covariance kernel function.
The Gaussian process condition is then satisfied if for any finite set of sample times $T=\{ t_1,\dots t_n\}$, the function evaluated at the times in $T$ presents samples from a multivariate Gaussian distribution. Let $d=1$, the mean and covariance functions, evaluated at $T$, be a vector and a matrix respectively. This multivariate Gaussian is denoted by
\begin{align*}
    F(T) \sim \mathcal{N}(M(T),K(T,T)).
\end{align*}

The predictive distribution for any test time, $t$, and zero prior mean is given by 
\begin{align}
    p(F(t)\, | \, D,t) &= \mathcal{N}(\mu(t),\sigma^2(t)),
        \label{eq:gp_prob}
    \\
    \mu(t) &= K(T,t) ^\top ( K(T,T) + \Sigma^2)^{-1} Z\nonumber
    \\
    \sigma^2(t) & = K(t,t) \nonumber
    \\
    &- K(T,t) ^\top( K(T,T) + \Sigma^2)^{-1}K(T,t),\nonumber
\end{align}
where $D$ is a collection of times and corresponding  output measurements $D=\{T,\,Z\}$. 
The measurement covariance matrix is denoted $\Sigma^2$. 
Given the dataset, the posterior mean and covariance, as shown in Equation~\eqref{eq:gp_prob}, are completely determined by the choice of covariance function, $K$.  The covariance function used in this paper is the Wiener velocity model, given by the equation:
\begin{align*}
    &K(t,t^{\top}) 
    \\
    &= \theta \left[\frac{1}{3} \textnormal{ min}^3({t},{t}^{\top}) +\frac{1}{2}\left|t-t^{\top}\right|\textnormal{ min}^2({t},{t}^{\top})\right],
\end{align*}
where $\theta$ is the length scale hyper-parameter. This covariance function is chosen because the trajectories produced are not overly smooth, and the function is non-stationary, which prevents the estimation from returning to the prior distribution. This particular covariance function will predict linearly outside the range of the observed data. Further, this covariance function admits a finite-state space model, which allows the prior mean and covariance to be estimated numerically efficiently using the methods demonstrated in~\cite{nickisch_state_2018}. 

\subsection{Metrics}
\subsubsection{Wasserstein Metric}
In this work, we use the Wasserstein metric~\cite{mallasto2017}, which will be used for multivariate Gaussian distributions. This metric is useful in this context, where we wish to compute the centroid of multiple multivariate Gaussian distributions. This metric, as used in this work, allows distributions to be combined while considering both the distribution means and covariances.
For  the multivariate Gaussian distribution used in this work, the Wasserstein metric is given by:
\begin{align}
    D_W(f_1,f_2)^2 &= d^2_2(m_1,m_2) \nonumber
    \\
    &+ \mathrm{Tr}\left(S_1 + S_2 - 2\left(S_1^{\frac{1}{2}} S_2 S_1^{\frac{1}{2}}\right) \right),
    \label{eq:wass}
\end{align}
where $d^2_2(\cdot,\cdot)$ is the Euclidean distance, $f_1$ and $f_2$ are multivariate Gaussian distributions with means $m_1$ and $m_2$, covariances $S_1$ and $S_2$, and $\mathrm{Tr}(\cdot)$ is the trace operator.

\subsubsection{Mahalanobis Metric}
This work additionally makes use of the Mahalanobis metric, which is used as a measure of dissimilarity of an observation from a distribution. The metric is defined for a multivariate Gaussian distribution as:
\begin{align}
    D_M(x,f) = \sqrt{(x-m)^\top S^{-1}(x-m)}, 
    \label{eq:mahal}
\end{align}
where $x$ is the observation being measured against the distribution $f$, which is a multivariate Gaussian distribution with mean $m$ and covariance $S$. This metric is used to compute the metric between the observed vehicle trajectory and the constructed intersection traffic model.

\subsection{Data Source}
\label{sec:data}
The data set is assumed to be constructed of sequential data. Each element in the sequence is minimally composed of a timestamp, an index indicating the trajectory to which the point belongs, and an $x$-$y$ position. This sequential data is assumed to be the output of an object recognition algorithm which has two known limitations. The first is that the data may not be uniformly sampled in time. This can occur if the online video feed freezes or if the object recognition algorithm fails to find the object at a specific time. The second limitation is that the recognition software is assumed to provide only the center of the vehicle's image. The center of the vehicle image is expected to change with orientation, which introduces error into the position measurement. 

\begin{figure}[t!]
  \centering
  \includegraphics[trim={0.75cm  0cm 0.75cm 0cm}, clip,width=\columnwidth]{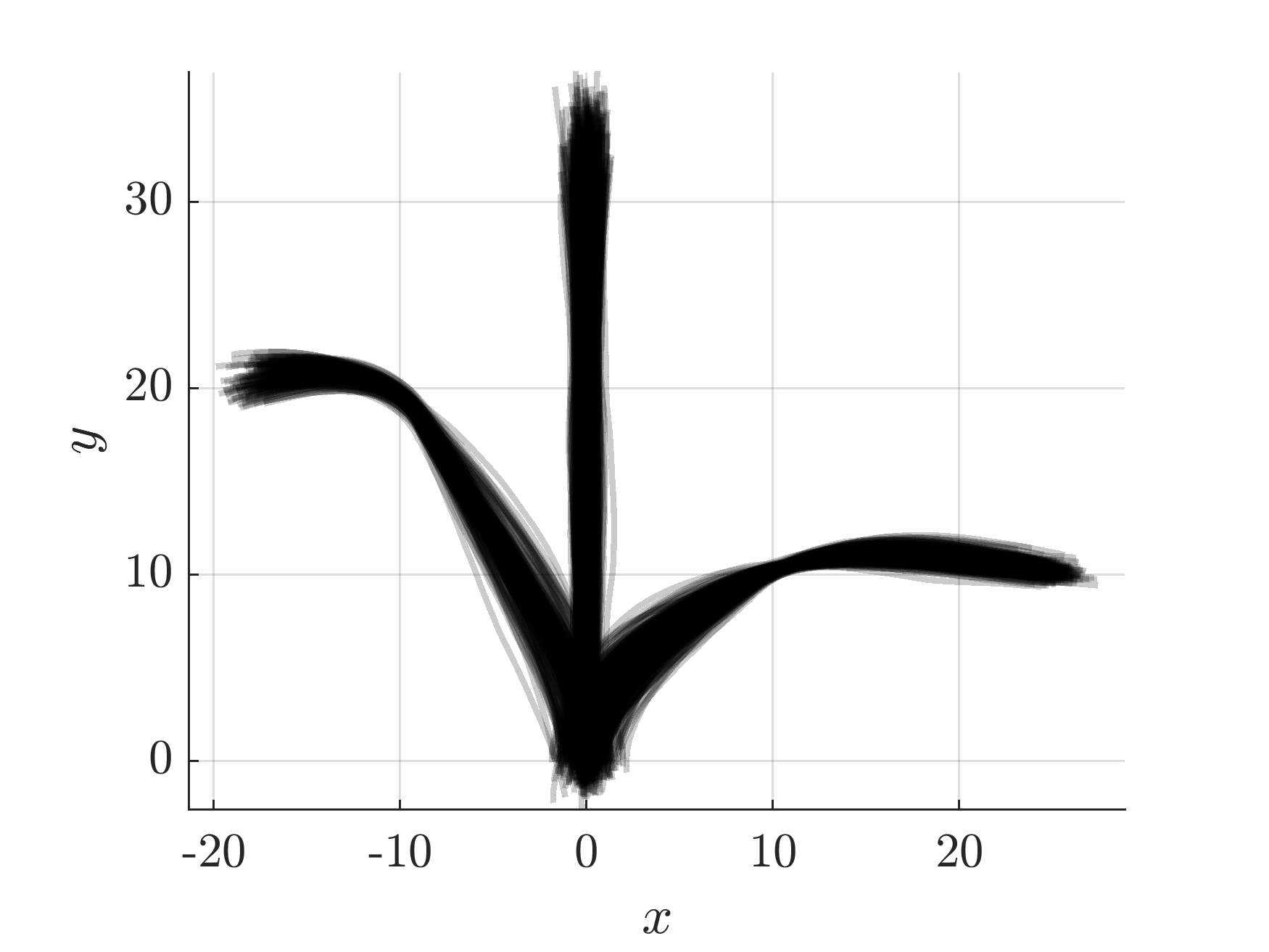}
  \caption{Time-series data from 1000 simulated trajectories in an intersection.}
  \label{fig:data}
\end{figure}

From this labeled sequence, we can re-order the data into a set of $x$-$y$ position trajectories sampled non-uniformly in time. The length of time over which these trajectories occur is not constant, each trajectory has a different time interval. The first step is to normalize the starting time to zero. Secondly, the lengths of time over which the trajectories occur are homogenized. This is performed by discarding trajectories that are either too long or too short. The remaining trajectories are truncated if they are too long, and the short trajectories are extrapolated. A visualization of  a thousand such trajectories is shown in Figure~\ref{fig:data}.
\section{Problem Formulation}
\label{sec:problem}
In this section we present the definitions used throughout the manuscript. Using these definitions, we formalize the  contributions with three problem statements. 
\subsection{Definitions}
\subsubsection{Intention}
We define the probabilistic intention, $I$, as a distribution over positions and velocities at a future time,  the intention time, $t_I$. The variables $x$ and $y$ indicate positions in the plane.
 For each dimension in the plane, indicated by subscript $x$ or $y$, we assume that the intention follows a normal distribution:
\begin{align*}
   & I_x \sim \mathcal{N}(\mu_{Ix},\sigma^2_{Ix}), \qquad I_y \sim \mathcal{N}(\mu_{Iy},\sigma^2_{Iy}),
\end{align*}
with associated means and variances.
\subsubsection{Intended Trajectory}
As described above, the intention of the vehicle is defined as a distribution at a fixed time. The intended trajectory indicates the set of intentions parameterized by time. Given this trajectory, the intention can be generated by sampling at the intention time, $t_I$. Recalling the definition of a Gaussian process, where any sample is a multivariate Gaussian distribution, we require the intended trajectory to be a Gaussian process.  

The intended trajectory is defined for each dimension in the plane, indicated by subscript $x$ or $y$. We denote the indented trajectories in the following equations:
\begin{align*}
    &\mathcal{I}_x(t) \sim \mathcal{GP}(M_x(t),K_x(t,t)), 
    \\ 
    &\mathcal{I}_y(t) \sim \mathcal{GP}(M_y(t),K_y(t,t)).
\end{align*}
The lack of subscript on the variable $t$ indicates that it is free to vary, unlike in the case of the intention where we have a fixed time $t_I$.

\subsubsection{Intersection Traffic Model}
The intersection traffic model is a set of indexed intended trajectories. In a given intersection, there may be many possible feasible vehicle trajectories. Though the vehicle may only follow one trajectory, it is assumed that the intended trajectory is chosen from a set of possibilities. This set of intended trajectories is modeled as a set of Gaussian process:
\begin{align}
    &\mathcal{I}_x(t,k) \sim \mathcal{GP}(M_x^k(t),K_x^k(t,t)),\nonumber
    \\
    &\mathcal{I}_y(t,k) \sim \mathcal{GP}(M_y^k(t),K_y^k(t,t)),
    \label{eq:set}
\end{align}
where the value of $k$ is the index of the intended trajectory.

From this set of intended trajectories, we can retrieve the intention or intended trajectory by fixing the values of $t$ and $k$. If we know the index of the intended trajectory of a vehicle, we fix the value of $k$ and retrieve a single Gaussian process, representing the intended trajectory of the vehicle. If we sample the Gaussian process at a specific time $t_I$, we recover the intention of the vehicle, which is a distribution over the vehicle's position. 

\subsubsection{Continuous Trajectory Set}
To learn the model for an intersection that can be sampled at any time, a set of continuous trajectories must be constructed. This set of functions, $\mathcal{E}$, takes values in $\mathbb{R}^J$ when evaluated at a single time, where $J$ is the number of trajectories. If evaluated with a vector of times, $T$, then the function produces a matrix of values $\mathbb{R}^{J \times N}$, where $N$ is the number of times.

\subsection{Problems}
The following problems are considered in this paper:
\subsubsection{Problem 1 - Trajectory Set Construction}
\label{prob1}
Given the non-uniformly sampled time-series trajectories described in Section~\ref{sec:data}, create a trajectory set $\mathcal{E}$ that can be sampled at any time.
\subsubsection{Problem 2 - Intersection Traffic Model}
Given the constructed dataset of trajectories, $\mathcal{E}$, construct a set of intended trajectories, described by Equation~\eqref{eq:set}, that captures the variance of all vehicles passing through the intersection.

\subsubsection{Problem 3 - Online Classification}
Given the constructed set of intended trajectories, ($\mathcal{I}_x(t,k)$ and $\mathcal{I}_x(t,k)$) and a dataset, $D$, of observed vehicle positions, determine which intended trajectory best models the behavior of the vehicle.

\section{Method}
\label{sec:method}
In this section, we present the process that generates the intersection traffic model and demonstrate an example usage of the model for classifying observed vehicle data.
The first step in this modeling procedure is to preprocess the data.

\subsection{Trajectory Reconstruction}
\label{sec:reconstruct}
\begin{figure*}[h]
    \centering
    \begin{subfigure}[t]{0.45\textwidth}
        \centering
        \includegraphics[trim={0cm  0cm 0.75cm 0cm}, clip,width=\textwidth]{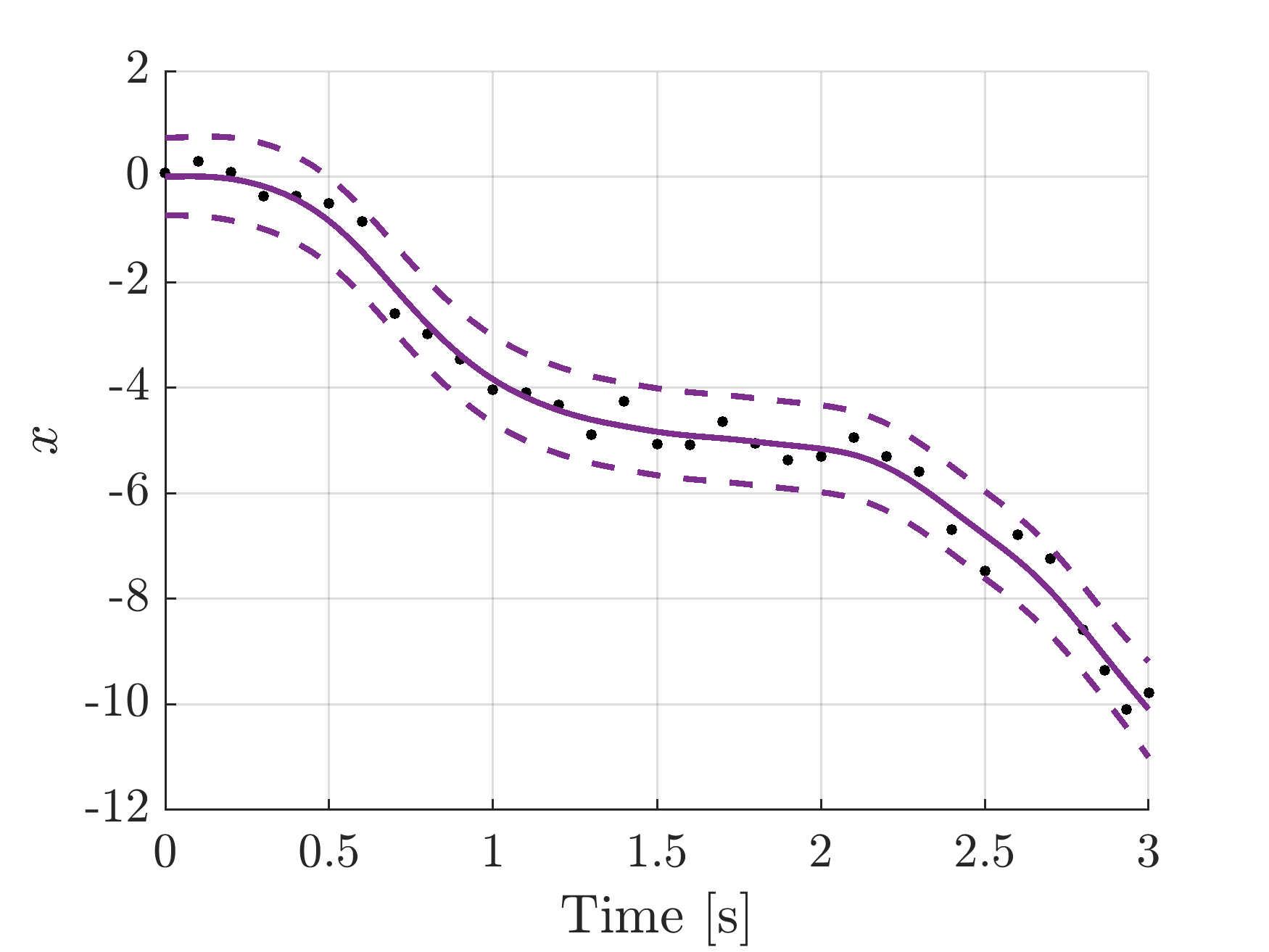}
        \caption{Reconstructed time-series trajectory for $x$ dimension.}
        \label{fig:2a}
    \end{subfigure}
    \hfill
    \begin{subfigure}[t]{0.45\textwidth}
        \centering
        \includegraphics[trim={0cm  0cm 0.75cm 0cm}, clip,width=\textwidth]{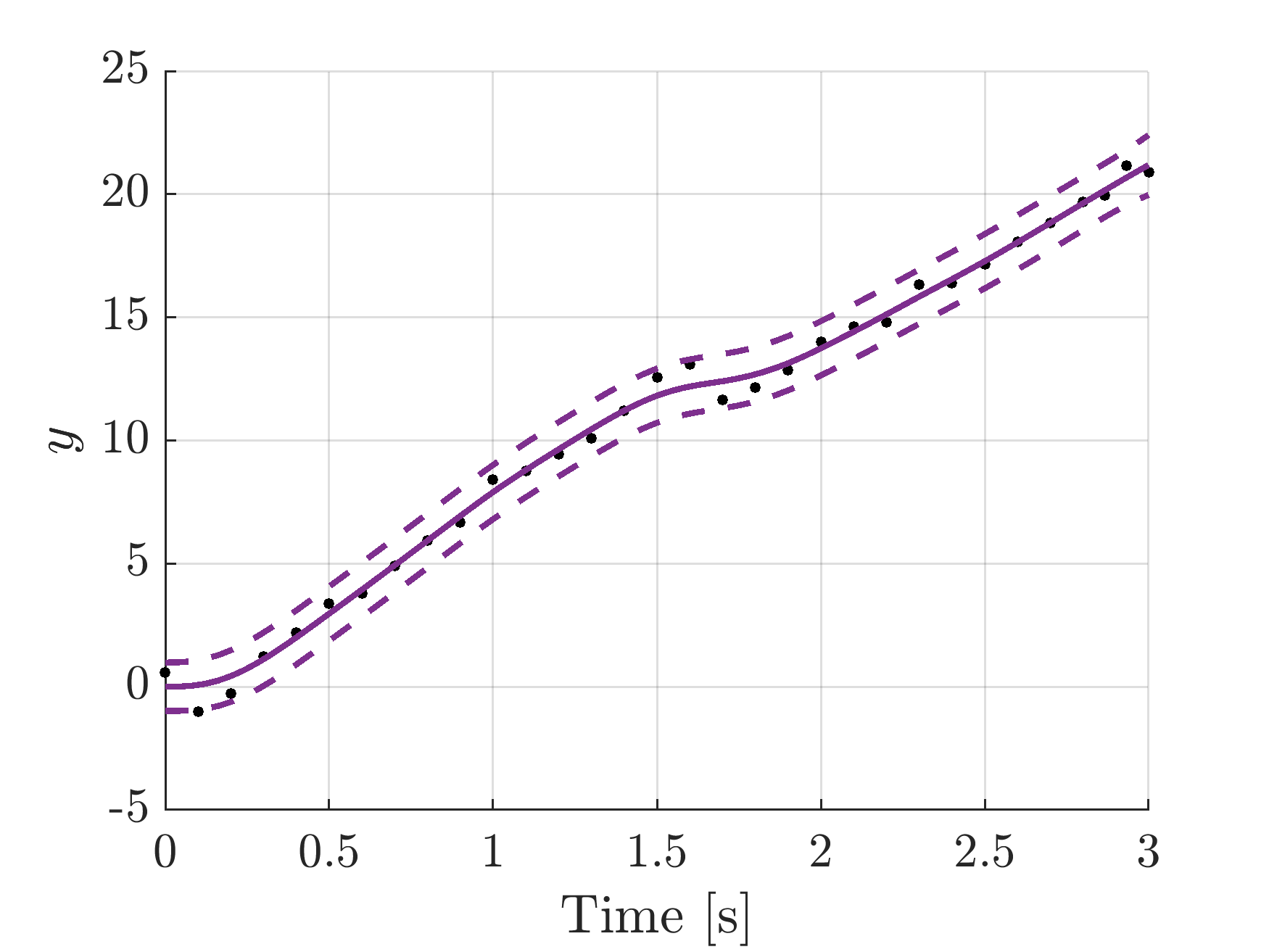}
        \caption{Reconstructed time-series trajectory for $y$ dimension.}
        \label{fig:2b}
    \end{subfigure}
        \caption{The output of the reconstruction procedure is shown for a single trajectory. The input data is shown as black points; the mean functions are shown as solid purple lines and the 2-$\sigma$ confidence intervals are indicated by the dotted lines.}
        \label{fig:recon}
\end{figure*}

The data used in this algorithm presents a non-uniformly sampled trajectory. To compare the data points at a given time, the trajectory values must be estimated at that time. 
This reconstruction can be performed using any interpolation method. In this work, we choose to use a Gaussian process formulation. This Gaussian process includes a mean function and a covariance function that satisfy the conditions of Problem~1.
Formally, for each observed trajectory indicated with subscript, $j$, the position observations, $Z_j$ and associated observation times, $T_j$, are collected into a dataset $D_j = \{T_j,Z_j\}$. Then by Equation~\eqref{eq:gp_prob}, we can write the estimated mean trajectory as
\begin{align}
    \mu_j(t) &= K(T_j,t) ^\top ( K(T_j,T_j) + \Sigma_j^2)^{-1} Z_j.
\end{align}
Similarly the posterior covariance can be computed.
To construct the trajectory dataset, we can collect these posterior means so that
\begin{align}
    \mathcal{E}(t) = \begin{bmatrix} \mu_1(t) \dots \mu_J(t) \end{bmatrix}^\top.
    \label{eq:trajset}
\end{align}
The results of this reconstruction procedure are shown for a single trajectory in Figure~\ref{fig:recon}. Note that the hyperparameters of each trajectory are optimized as described in~\cite{rasmussen06}.

\subsection{Trajectory Clustering}
Before the intersection traffic model is constructed, the  trajectories are first separated based on the cluster to which they belong. This operation is performed on the reconstructed datasets since the data can be compared at arbitrary times by sampling the trajectory dataset, $\mathcal{E}$. 

\begin{figure}[t]
  \centering
  \includegraphics[trim={0cm  0cm 0.75cm 0cm}, clip,width=\columnwidth]{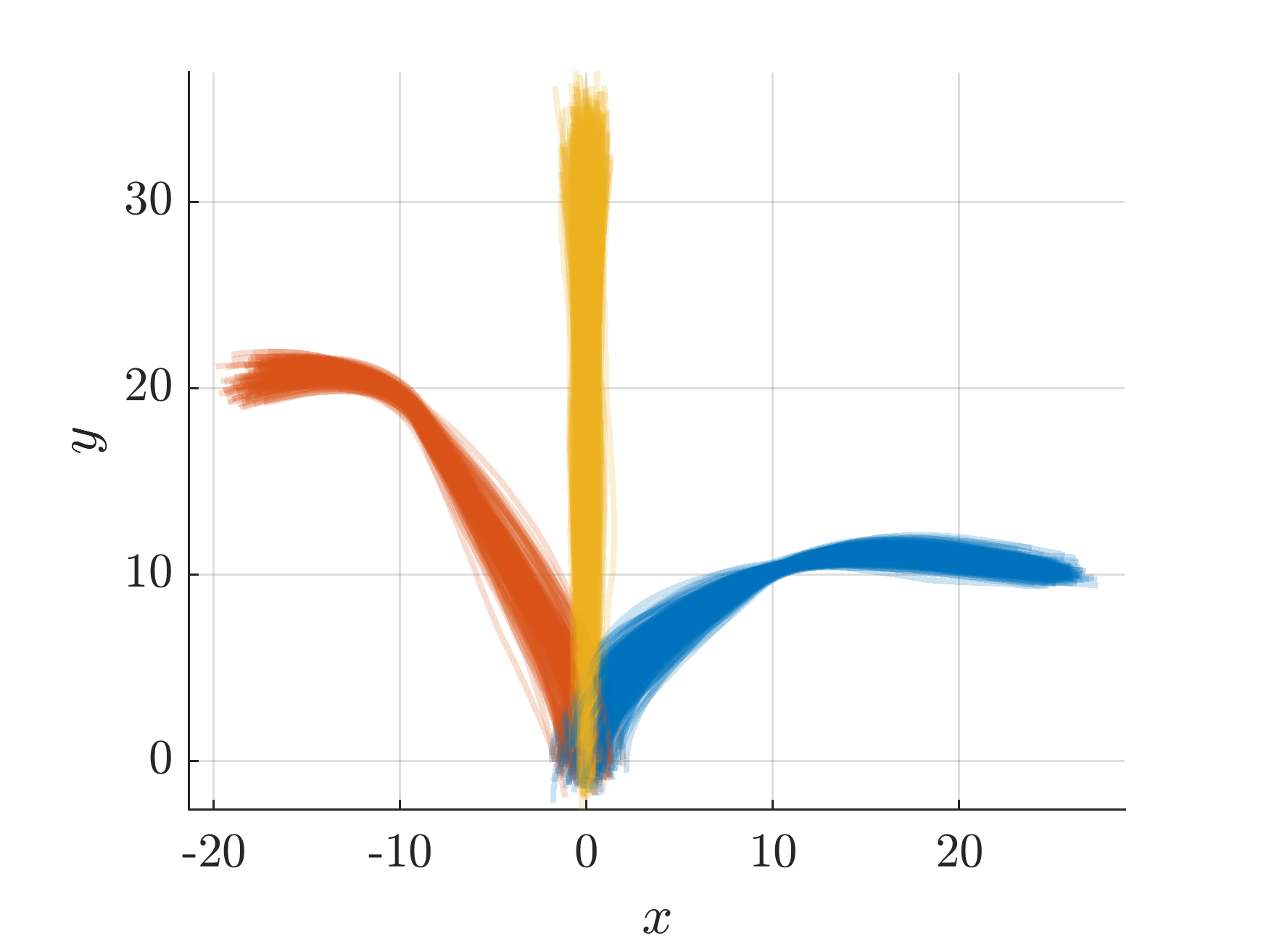}
  \caption{The output of the classification procedure is a label for each historical trajectory. Here the labels are used to color the raw data.}
  \label{fig:cluster}
\end{figure}

The number of clusters is chosen based on the design of the intersection. For the intersection considered, there is one source lane and there are 3 terminal lanes where the cars can drive. This leads to the possibility of three clusters.  The clusters are created using the \texttt{k-means++} algorithm~\cite{kmeans}, generating $K=3$ cluster labels by minimizing the Euclidean distance between the points and the assigned cluster center. The K-means algorithm is performed on the reconstructed data, sampled at a terminal time and the initial time, i.e. $\mathcal{E}([t_1,t_N])$, where $t_1=0$ is the initial time in the reconstructed time vector and $t_N=3$ is the final time.
The output of the clustering algorithm is a list of labels for each of the observed trajectories. An example of clustered data is shown in Figure~\ref{fig:cluster}, where colors are applied to raw data based on their cluster.

\subsection{Intersection Traffic Model}
The intersection traffic model is a set of intended trajectories. Each of these intended trajectories is constructed from the subset of observed trajectories based on the cluster. In this work, the number of clusters is $K=3$, therefore three intended trajectories are constructed. 

Consider the right-turn case, with $k=1$. For online use, we wish to pre-compute a set of points at which we expect the need to compare trajectories. The discretization times are contained in the vector $T$ and are chosen based on an assumed 20~Hz camera frame rate. Therefore the discretized time vector $T$ is a vector of $N=60$ data points uniformly spaced between 0 and 3~seconds. These time points are evaluated using the set of functions $\mathcal{E}(T)$, resulting in a matrix of dimension $J\times N$. Recall that $J=1000$ and $N=60$.

The mean, $M^k(T)$, is computed at these points by computing the empirical mean along the first dimension, producing value at each of the sample times.
Similarly, the empirical variance is generated by computing the outer product of the sampled trajectories minus their means. Recall that the sampled trajectories are the rows of $\mathcal{E}$, calculated from the functions $\mu_j$ as seen in Equation~\eqref{eq:trajset}. The variance estimate is given by the equation for each time:
\begin{align*}
    K^k(T,T) = \frac{1}{N-1}\sum_{n=1}^N (\mu_j-M^k)(\mu_j-M^k)^\top,
\end{align*}
where $N$ is the length of $T$. This procedure is repeated for  each cluster, in both the $x$ and $y$ dimensions.

\subsection{Trajectory Classification}
To classify an observed trajectory, we need to compare the observed data points to the intersection traffic model. The intersection traffic model is therefore produced as described above at times when we expect to observe data. If the observed data is missing, it is interpolated using the trajectory reconstruction method found in Section~\ref{sec:reconstruct}. This method ensures that the discretized model can be directly compared to the observed data.

To determine which cluster our currently observed vehicle belongs to, the unfinished trajectory must be classified. The trajectory is unfinished, meaning we only have data over a small portion of the intersection with which to decide if the vehicle belongs to a cluster. For that reason, the distance and threshold methods are chosen to allow recursive computation and allow the data to be compared directly to the intersection model.

The classification procedure is based on the Mahalanobis distance, shown in Equation~\eqref{eq:mahal}, which  compares our sampled $x$ and $y$ dimensional data to the means of each cluster with the covariance relationship $S$, being given by the intersection traffic model. At each observation time, the Mahalanobis distance is computed between the reconstructed observed trajectory and each of the sampled intended trajectories. For the full set of observation times $T$, and only considering the $x$ dimension, the distance to the first intended trajectory can be computed by substituting the definition of the intended trajectory, from Equation~\eqref{eq:set},
\begin{align*}
    x &\leftarrow \mu_x^O(T),\quad
    m \leftarrow M_x^1(T),\quad
    S \leftarrow K_x^1(T,T),
\end{align*}
into the Mahalanobis distance equation:
\begin{align*}
    D_M(x,f) &= \sqrt{(x-m)^\top S^{-1}(x-m)}.
\end{align*}
Here, $\mu_x^O(T)$ is the mean of the reconstructed observed trajectory evaluated at the times in the vector $T$.
This distance is calculated for both dimensions and added. The calculation is repeated for each cluster.

\begin{figure}[t]
  \centering
  \includegraphics[trim={0cm  0cm 0.75cm 0cm}, clip,width=\columnwidth]{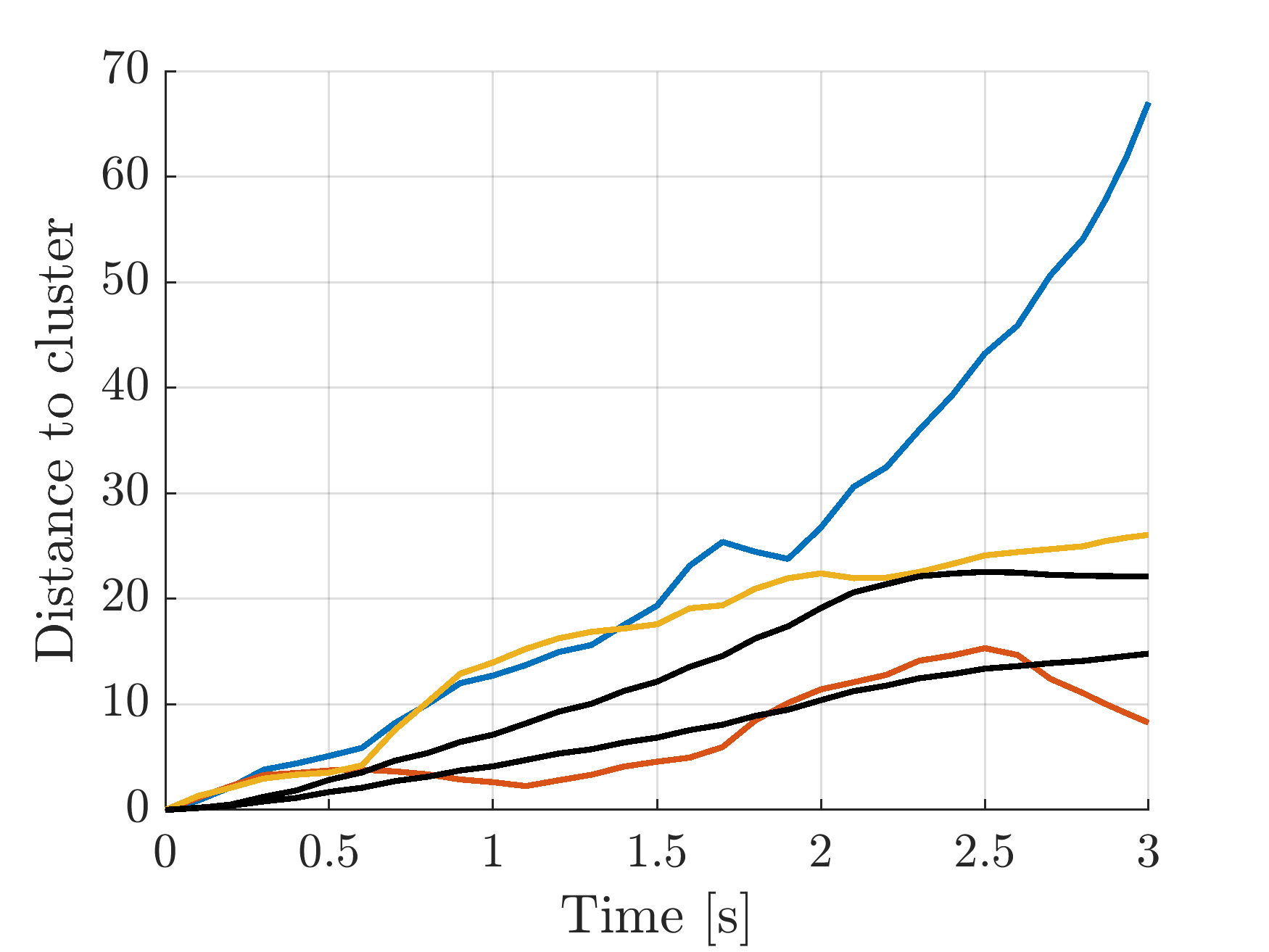}
  \caption{The distances from the observed data to each of the intended trajectories are shown as colored lines. The line colors correspond to the cluster. The black lines are the two exclusion thresholds. The yellow line remains above the two thresholds and is therefore excluded from consideration. The minimum distance is, therefore, the red cluster, and we would classify the observed data in cluster 2.}
  \label{fig:threshold}
\end{figure}

From this distance calculation, a simple decision threshold for classification would be to classify the trajectory as a member of the cluster that has minimum Mahalanobis distance to the mean of the observed data. In this work, we include an additional requirement.
The minimum Mahalanobis distance can be used for classification but only if the third cluster, the proceed straight trajectory, is ruled out. This additional criterion is added to demonstrate a possible method of constructing a threshold. 
This method uses an algorithm provided in~\cite{mallasto2017} that finds the centroid of multivariate Gaussian distributions using the Wasserstein metric introduced in Equation~\eqref{eq:wass}.
Using this method, we construct two new multivariate Gaussian distributions, one of which is equidistant between the left and center paths, and the other that is equidistant between the right and center paths. 
These distributions are then used to generate the exclusion threshold for the straight line threshold. 
The distance between the observed trajectory and the threshold distributions is computed as described above for the clusters. In this classification procedure, the default assumption is that the straight trajectory is the correct choice. However, if the observed mean  is farther from the center path than either of the thresholds, the minimum distance trajectory is selected. These distances are shown in Figure~\ref{fig:threshold}.
\section{Results}
\label{sec:results}
To evaluate the proposed method, a simulated dataset of 1000 trajectories from a single intersection is used. The cars considered in this intersection all start from the same lane and will either turn left, proceed straight or turn right. The raw data can be seen in Figure~\ref{fig:data}. This data is preprocessed and clustered in preparation for the creation of an intersection traffic model. The dataset is clustered with $K=3$. For each of these clusters, an intended trajectory is constructed. Together, these intended trajectories make up the intersection model.

To validate the modeling procedure, out-of-sample tests are conducted using a separate set of 1000 trajectories. For the out-of-sample tests, the held out trajectories are not used for any part of the modeling procedure.
All computations are performed in Windows using MATLAB version 2019b. The processor used is an Intel Core i5-4460 CPU operating at 3.2GHz, with 16 GB of RAM.

\subsection{Trajectory Clustering}
The trajectory clustering is performed on the training dataset shown in Figure~\ref{fig:data}. This clustering procedure produces labels, which are attached to each trajectory based on a sub-sampled trajectory. The clustered trajectories are shown in Figure~\ref{fig:cluster}.
\begin{figure}[t]
  \centering
  \includegraphics[trim={0cm  0cm 0.75cm 0cm}, clip,width=\columnwidth]{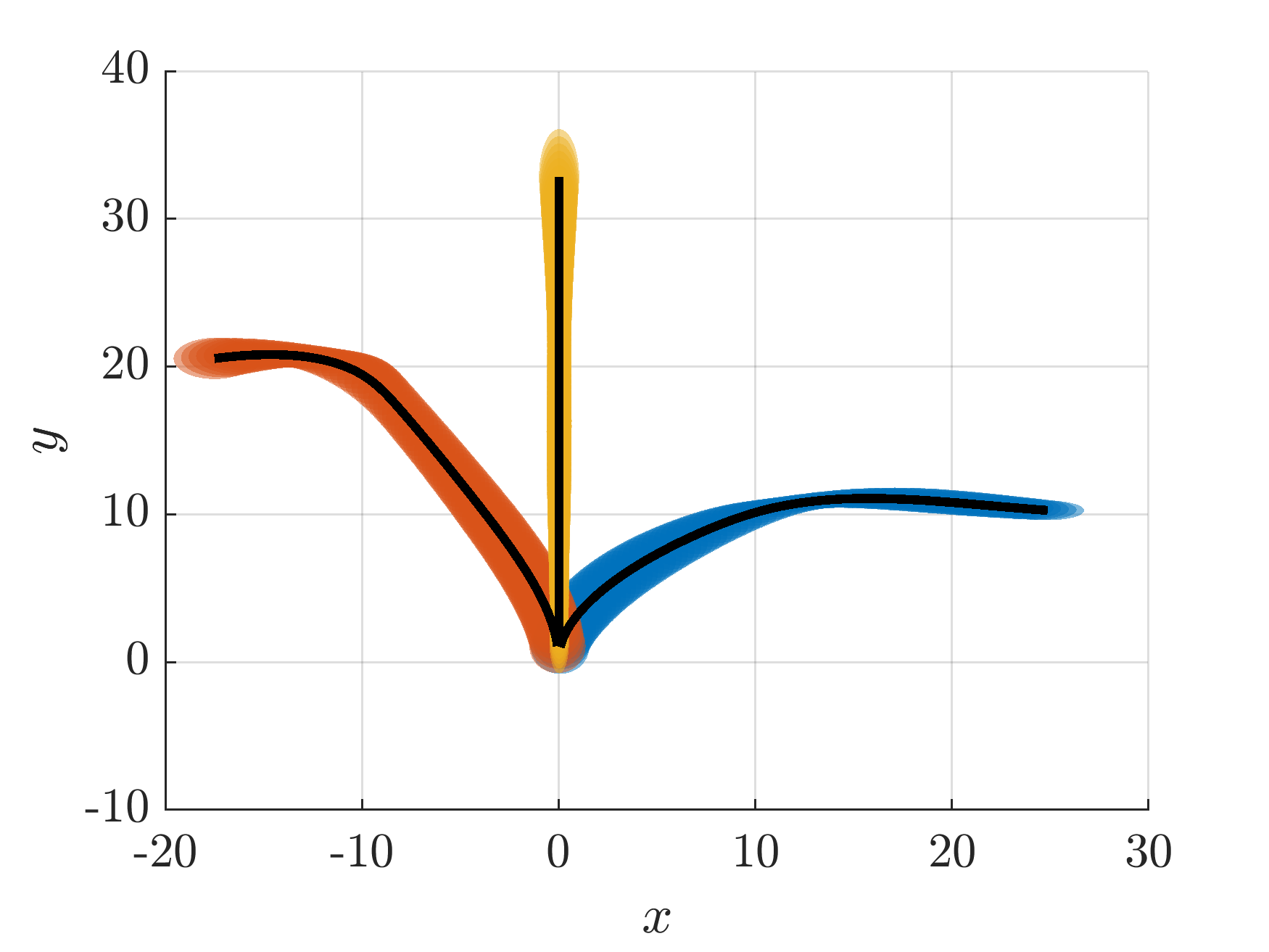}
  \caption{The intersection model is visualized, with different colors representing the  index of the intended trajectories. The black lines show the mean trajectory of each intended trajectory. The colored regions are the 2~$\sigma$ confidence ellipses, plotted at the sample times where the model is constructed. }
  \label{fig:model}
\end{figure}
We see that the clustering procedure produces the correct classification for all vehicles in this dataset. The time needed to cluster the data is shown in Table~\ref{tab:time}. The clustering procedure is completed in only 2~seconds.

\subsection{Intersection Traffic Model}
With the trajectories clustered, the Gaussian model can be created to capture the uncertainty in the intersection traffic model. For each cluster, the mean function is calculated as the sum of the posterior means of the trajectories with the same label. These mean functions are shown as solid black lines in Figure~\ref{fig:model}.
The variances are not shown, but the two standard deviation confidence regions are shown as colored regions in Figure~\ref{fig:model}. The computation time required to perform this modeling operation is shown in Table~\ref{tab:time}. The intersection model is constructed in approximately 5~seconds.

\subsection{Online Results}
The online results are computed using only partial information of a vehicle trajectory. The results are shown in Table~\ref{tab:time} to indicate the amount of computation time required to run for each new observation.

\subsubsection{Reconstruction}
As in the offline methods, Table~\ref{tab:time} shows that reconstruction takes up a significant amount of time. In this demonstration, the calculation time is approximately 0.04~seconds, which is sufficient for real-time operation at 20~Hz.

\subsubsection{Classification}
\begin{figure*}
    \centering
    \begin{subfigure}[t]{0.3\textwidth}
        \centering
        \includegraphics[trim={0cm  0cm 0.75cm 0cm}, clip,width=\textwidth]{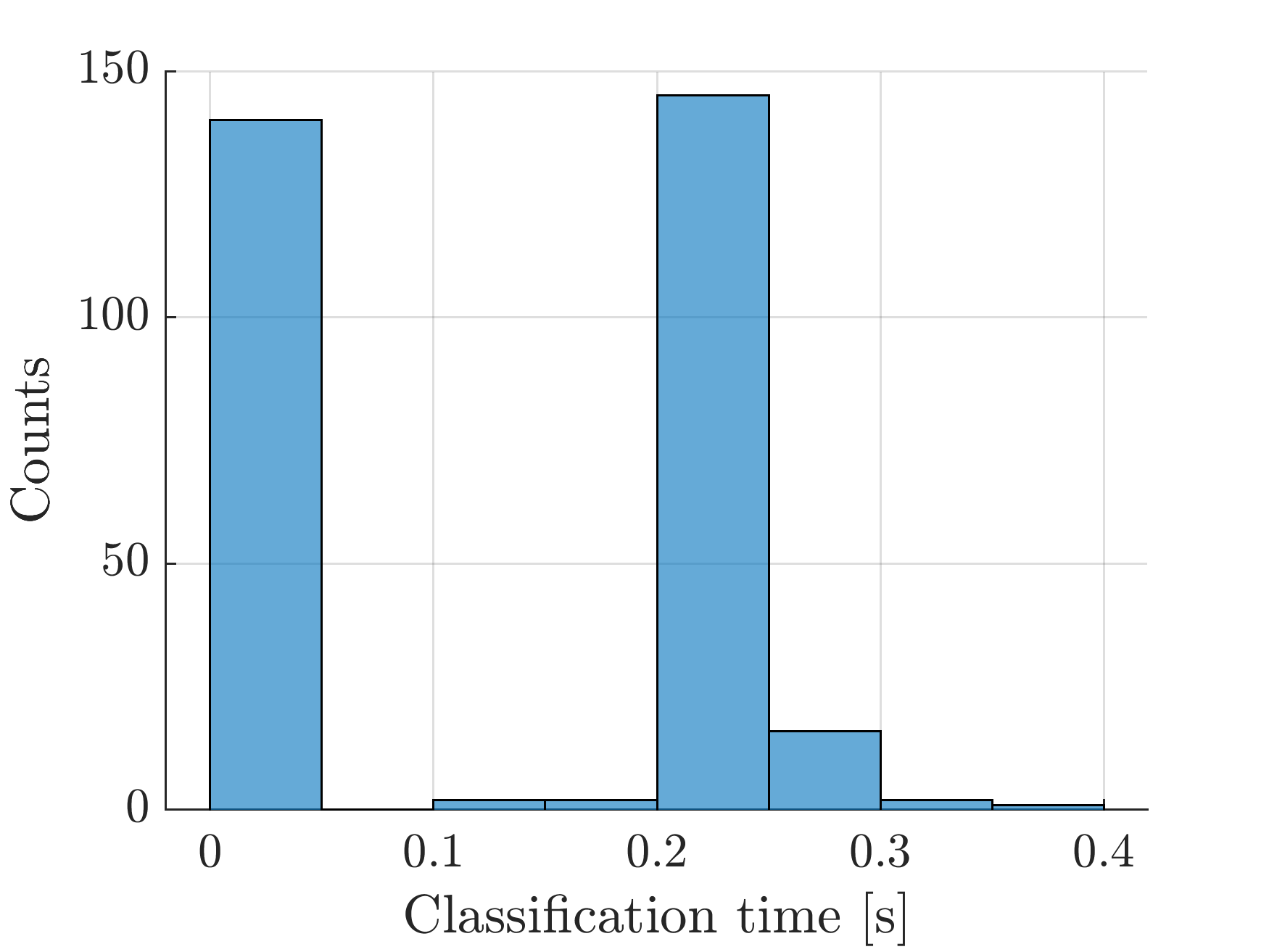}
        \caption{Cluster 1 - Right Turn}
        \label{fig:7a}
    \end{subfigure}
    ~
    \begin{subfigure}[t]{0.3\textwidth}
        \centering
        \includegraphics[trim={0cm  0cm 0.75cm 0cm}, clip,width=\textwidth]{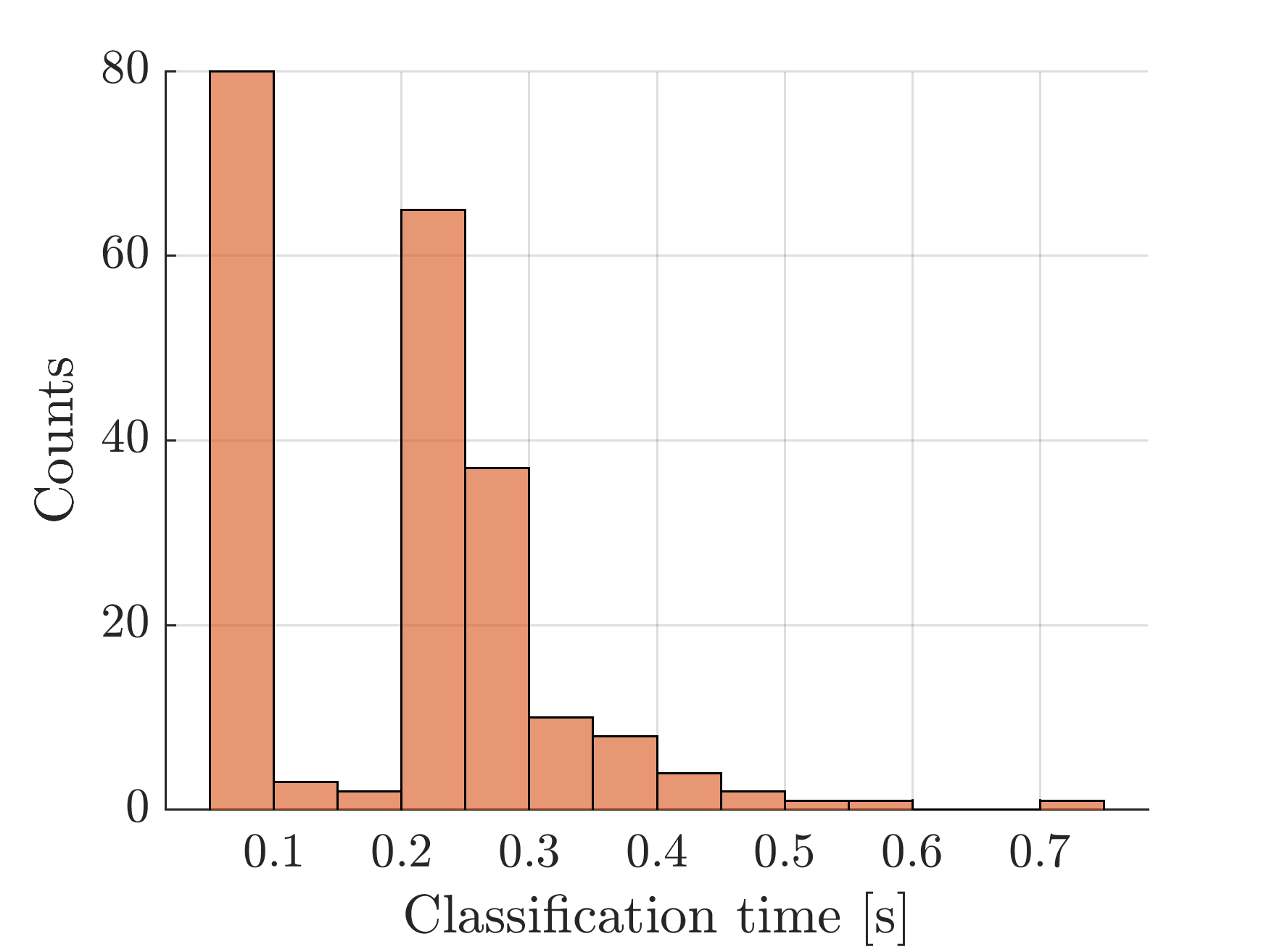}
        \caption{Cluster 2 - Left Turn}
        \label{fig:7b}
    \end{subfigure}
    ~
    \begin{subfigure}[t]{0.3\textwidth}
        \centering
        \includegraphics[trim={0cm  0cm 0.75cm 0cm}, clip,width=\textwidth]{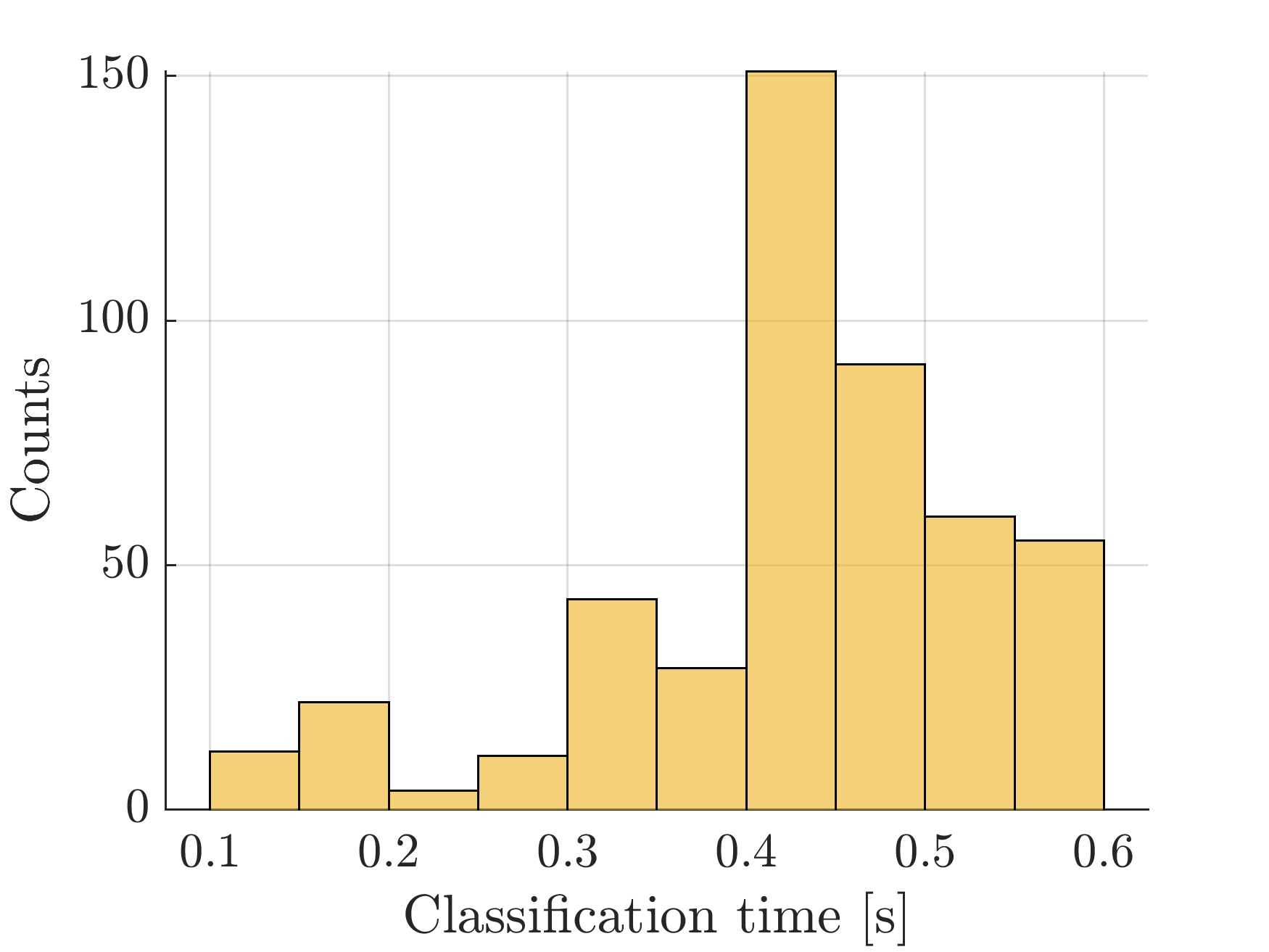}
        \caption{Cluster 3 - Straight}
        \label{fig:7c}
    \end{subfigure}
    \caption{Histogram of classification times. This is the time at which the algorithm produces the correct cluster as a guess and holds that guess for the rest of the scenario.}
  \label{fig:hists}
\end{figure*}

In Figure~\ref{fig:threshold} we see that given partial observations of a car turning left, the cluster distances change over time. These distances are then used for classification. The distances to the clusters are shown as colored lines and the black lines show the exclusion thresholds. Recall that the output of the classification procedure at each time is the index of the cluster with the minimum distance to the observed data, as long as the yellow line is greater than the exclusion thresholds. Using this criterion, we see that the correct classification occurs after $0.6$~s of observed data, after which the distance to the left turn cluster is smaller than the other clusters. The computation time required to perform classification at each time step is shown in Table~\ref{tab:time}. 
Given the full trajectories, 100\% of the online estimates converged to the correct classification. 
Ideally, the classification would occur before the trajectory was finished. To quantify the performance of the classification procedure, the distributions of classification times are shown in Figure~\ref{fig:hists}. The classification time is the amount of observation time it takes for the algorithm to correctly guess the cluster to which the trajectory belongs. For example, given validation trajectories from cluster 1, indicated in blue in Figure~\ref{fig:7a}, the algorithm correctly guessed that the trajectory belonged to cluster 1 and held that guess for the rest of the 3~second time interval 50\% of the time. In Figure~\ref{fig:7c} we see that the straight-line trajectory was not finally identified until on average about 0.5~seconds. This time interval can be compared to the reconstructed trajectory in Figure~\ref{fig:recon}. Even the worst classification times still identify the correct cluster before the observed vehicle has moved significantly.

\begin{table}
  \caption{Computation Times}
  \label{tab:time}
  \centering
  \begin{tabular}{llc}
    \toprule
    Offline Algorithm     &  Computation Time  [s]  & \texttt{Online?} \\
    \midrule
    Dataset Reconstruction      & 835               &  \ding{55}\\
    Dataset Clustering          & 2                 &  \ding{55}\\
    Intersection Traffic Modeling       & 5                 &  \ding{55} \\
    Trajectory Reconstruction   & 0.04              &  \ding{51} \\
    Trajectory Classification   & 0.0003            &  \ding{51} \\
    \bottomrule
  \end{tabular}
\end{table}

\section{Conclusion}
\label{sec:conclusion}
In this work, we have demonstrated the construction of a probabilistic traffic model for a vehicle intersection from a set of position measurements. The model is constructed from a set of 1000 trajectories and is constructed in under 15~minutes. 
This method's applicability to vehicle trajectory prediction is demonstrated in a classification task where a vehicle trajectory is  reconstructed and classified as a left-turn, right-turn or straight trajectory. Both reconstruction and classification are  completed in approximately 40~ms, making them suitable for real-time applications.

Though this work focuses on intersection modeling, this method may apply to other scenarios such as traffic merging and lane changing. Future work includes extending the presented method to these scenarios to create a more complete model of traffic flow. In addition to different scenarios, for a complete treatment of this traffic modeling one needs to consider vehicle-vehicle interaction. Future work will not only create a model from observed trajectories but also learn from interactions that occur when multiple cars are observed in the same environment.

\section*{Acknowledgments}
This work is supported by the National Aeronautics and Space Administration (NASA). 

\bibliographystyle{plainnat}
\bibliography{references}

\begin{thebibliography}{20}
\providecommand{\natexlab}[1]{#1}
\providecommand{\url}[1]{\texttt{#1}}
\expandafter\ifx\csname urlstyle\endcsname\relax
  \providecommand{\doi}[1]{doi: #1}\else
  \providecommand{\doi}{doi: \begingroup \urlstyle{rm}\Url}\fi

\bibitem[Arthur and Vassilvitskii(2007)]{kmeans}
D.~Arthur and S.~Vassilvitskii.
\newblock \href{https://dl.acm.org/doi/10.5555/1283383.1283494} {K-means++: The
  Advantages of Careful Seeding}.
\newblock In \emph{Proceedings of the 18th ACM-SIAM Symposium on Discrete
  Algorithms}, pages 1027–--1035, Philadelphia, PA, USA, 2007.

\bibitem[Barratt et~al.(2019)Barratt, Kochenderfer, and Boyd]{barratt2019}
S.~Barratt, M.~Kochenderfer, and S.~Boyd.
\newblock \href{https://ieeexplore.ieee.org/document/8551278}{Learning
  Probabilistic Trajectory Models of Aircraft in Terminal Airspace From
  Position Data}.
\newblock \emph{IEEE Transactions on Intelligent Transportation Systems},
  20\penalty0 (9):\penalty0 3536--3545, 2019.

\bibitem[Cichella et~al.(2017)Cichella, Marinho, Stipanović, Hovakimyan,
  Kaminer, and Trujillo]{cichella17}
V.~Cichella, T.~Marinho, D.~Stipanović, N.~Hovakimyan, I.~Kaminer, and
  A.~Trujillo.
\newblock
  \href{https://link.springer.com/article/10.1007/s10846-017-0517-6}{Collision
  Avoidance Based on Line-of-Sight Angle: Guaranteed Safety Using Limited
  Information About the Obstacle}.
\newblock \emph{Journal of Intelligent Robot Systems}, 89\penalty0
  (2):\penalty0 139--153, 2017.

\bibitem[Kaminer et~al.(2017)Kaminer, Pascoal, Xargay, Hovakimyan, Cichella,
  and Dobrokhodov]{kaminer17}
I.~Kaminer, M.~Pascoal, E.~Xargay, N.~Hovakimyan, V.~Cichella, and
  V.~Dobrokhodov.
\newblock
  \emph{\href{https://www.elsevier.com/books/time-critical-cooperative-control-of-autonomous-air-vehicles/kaminer/978-0-12-809946-9}{Time-Critical
  Cooperative Control of Autonomous Air Vehicles}}.
\newblock Elsevier, 2017.

\bibitem[Mallasto and Feragen(2017)]{mallasto2017}
A.~Mallasto and A.~Feragen.
\newblock
  \href{https://papers.nips.cc/paper/7149-learning-from-uncertain-curves-the-2-wasserstein-metric-for-gaussian-processes}{Learning
  from Uncertain Curves: The 2-{W}asserstein Metric for {G}aussian Processes}.
\newblock In \emph{Proceedings of the 31st International Conference on Neural
  Information Processing}, pages 5665–--5674, Long Beach, CA, USA, 2017.

\bibitem[Marinho et~al.(2018)Marinho, Amrouche, Cichella, Stipanović, and
  Hovakimyan]{marinho18}
T.~Marinho, M.~Amrouche, V.~Cichella, D.~Stipanović, and N.~Hovakimyan.
\newblock \href{https://ieeexplore.ieee.org/document/8431871}{Guaranteed
  Collision Avoidance Based on Line-of-Sight Angle and Time-to-Collision}.
\newblock In \emph{Proceedings of American Control Conference}, pages
  4305–--4310, Milwaukee, WI, USA, 2018.

\bibitem[Mehdi et~al.(2019)Mehdi, Choe, and Hovakimyan]{bilal19}
S.~Mehdi, R.~Choe, and N.~Hovakimyan.
\newblock \href{https://arc.aiaa.org/doi/full/10.2514/1.G003864}{Collision
  Avoidance in Cooperative Missions: Bézier Surfaces for Circumnavigating
  Uncertain Speed Profiles}.
\newblock \emph{AIAA Journal of Guidance, Control, and Dynamics}, 42\penalty0
  (8):\penalty0 1779--–1796, 2019.

\bibitem[Morton et~al.(2017)Morton, Wheeler, and
  Kochenderfer]{morton_analysis_2017}
J.~Morton, T.~Wheeler, and M.~Kochenderfer.
\newblock \href{https://ieeexplore.ieee.org/abstract/document/7565491}{Analysis
  of Recurrent Neural Networks for Probabilistic Modeling of Driver Behavior}.
\newblock \emph{IEEE Transactions on Intelligent Transportation Systems},
  18\penalty0 (5):\penalty0 1289--–1298, 2017.

\bibitem[Nickisch et~al.(2018)Nickisch, Solin, and
  Grigorievskiy]{nickisch_state_2018}
H.~Nickisch, A.~Solin, and A.~Grigorievskiy.
\newblock \href{http://proceedings.mlr.press/v80/nickisch18a.html}{State Space
  {G}aussian Processes with Non-{G}aussian Likelihood}.
\newblock In \emph{Proceedings of the 35th International Conference on Machine
  Learning}, pages 3789–--3798, Stockholm, Sweden, 2018.

\bibitem[Paden et~al.(2016)Paden, Čáp, Yong, Yershov, and Frazzoli]{paden16}
B.~Paden, M.~Čáp, S.~Yong, D.~Yershov, and E.~Frazzoli.
\newblock \href{https://ieeexplore.ieee.org/abstract/document/7490340}{A Survey
  of Motion Planning and Control Techniques for Self-driving Urban Vehicles}.
\newblock \emph{IEEE Transactions on Intelligent Vehicles}, 1\penalty0
  (1):\penalty0 33--55, 2016.

\bibitem[Pan et~al.(2019)Pan, Sun, Xu, Jiang, Xiao, Hu, and
  Miao]{pan_lane_2019}
J.~Pan, H.~Sun, K.~Xu, Y.~Jiang, X.~Xiao, J.~Hu, and J.~Miao.
\newblock \href{https://arxiv.org/abs/1909.13377}{Lane Attention: Predicting
  Vehicles' Moving Trajectories by Learning Their Attention over Lanes}.
\newblock \emph{arXiv:1909.13377}, 2019.

\bibitem[Patterson et~al.(2019)Patterson, Lakshmanan, and
  Hovakimyan]{patterson2019}
A.~Patterson, A.~Lakshmanan, and N.~Hovakimyan.
\newblock \href{https://arxiv.org/abs/1904.02765}{Intent-Aware Probabilistic
  Trajectory Estimation for Collision Prediction with Uncertainty
  Quantification}.
\newblock In \emph{Proceedings of 58th IEEE Conference on Decision and
  Control}, pages 3827--3832, Nice, France, 2019.

\bibitem[Rasmussen and Williams(2006)]{rasmussen06}
C.~Rasmussen and C.~Williams.
\newblock \emph{\href{http://www.gaussianprocess.org/gpml/}{Gaussian Processes
  for Machine Learning}}.
\newblock The MIT Press, 2006.

\bibitem[Ren et~al.(2018)Ren, Wang, Laskey, and Goldberg]{ren2016}
X.~Ren, D.~Wang, M.~Laskey, and K.~Goldberg.
\newblock \href{https://ieeexplore.ieee.org/document/8560597}{Learning Traffic
  Behaviors by Extracting Vehicle Trajectories from Online Video Streams}.
\newblock In \emph{Proceedings of 14th IEEE International Conference on
  Automation Science and Engineering}, pages 1276--1283, Vancouver, BC, Canada,
  2018.

\bibitem[Roberts et~al.(2013)Roberts, Osborne, Ebden, Reece, Gibson, and
  Aigrain]{roberts12}
S.~Roberts, M.~Osborne, M.~Ebden, S.~Reece, N.~Gibson, and S.~Aigrain.
\newblock
  \href{https://royalsocietypublishing.org/doi/full/10.1098/rsta.2011.0550}{Gaussian
  Processes for Time-Series Modelling}.
\newblock \emph{Philosophical Transactions of the Royal Society A:
  Mathematical, Physical and Engineering Sciences}, 371\penalty0
  (1984):\penalty0 20110550, 2013.

\bibitem[Wurts et~al.(2018)Wurts, Stein, and Ersal]{wurts18}
J.~Wurts, J.~Stein, and T.~Ersal.
\newblock \href{https://ieeexplore.ieee.org/document/8431536}{Collision
  Imminent Steering using Nonlinear Model Predictive Control}.
\newblock In \emph{Proceedings of American Control Conference}, pages
  4772--4777, Milwaukee, WI, USA, 2018.

\bibitem[Xin et~al.(2018)Xin, Wang, Chan, Chen, Li, and
  Cheng]{xin_intention-aware_2018}
L.~Xin, P.~Wang, C.~Chan, J.~Chen, S.~Li, and B.~Cheng.
\newblock \href{https://ieeexplore.ieee.org/document/8569595}{Intention-aware
  Long Horizon Trajectory Prediction of Surrounding Vehicles using Dual {LSTM}
  Networks}.
\newblock In \emph{Proceedings of 21st IEEE International Conference on
  Intelligent Transportation Systems}, pages 1441--1446, Maui, HI, USA, 2018.

\bibitem[Yokoyama(AIAA 2018-1857, 2018)]{yokoyama18}
N.~Yokoyama.
\newblock \href{https://arc.aiaa.org/doi/abs/10.2514/6.2018-1857}{Decentralized
  Conflict Detection and Resolution using Intent-Based Probabilistic Trajectory
  Prediction}.
\newblock In \emph{Proceedings of AIAA Guidance, Navigation, and Control
  Conference}, Kissimmee, FL, USA, AIAA 2018-1857, 2018.

\bibitem[Yoon et~al.(2019)Yoon, Widdowson, Marinho, Wang, and
  Hovakimyan]{yoon_socially_2019}
H.~Yoon, C.~Widdowson, T.~Marinho, R.~Wang, and N.~Hovakimyan.
\newblock \href{https://dl.acm.org/doi/10.1145/3341570}{Socially Aware Path
  Planning for a Flying Robot in Close Proximity of Humans}.
\newblock \emph{ACM Transactions on Cyber-Physical Systems}, 3\penalty0
  (4):\penalty0 41:1--41:24, 2019.

\bibitem[Zhu et~al.(2019)Zhu, Qin, Wang, and Zhao]{zhu_probabilistic_2019}
J.~Zhu, S.~Qin, W.~Wang, and D.~Zhao.
\newblock \href{https://arxiv.org/abs/1910.08102}{Probabilistic Trajectory
  Prediction for Autonomous Vehicles with Attentive Recurrent Neural Process}.
\newblock \emph{arXiv:1910.08102}, 2019.

\end{thebibliography}

\end{document}